\documentclass[lettersize,journal]{IEEEtran}

\usepackage[utf8]{inputenc}
\usepackage{wrapfig}
\usepackage[normalem]{ulem}
\usepackage{nicefrac}
\usepackage{mathptmx}                  
\usepackage{amsfonts}
\usepackage{amsmath}
\usepackage{tikz}
\usepackage{href-ul}
\usepackage{orcidlink}
\usepackage[mathcal]{eucal} 
\let\rm=\rmfamily

\def \N {\mbox{\rm \hbox{I\kern-.15em\hbox{N}}}}
\def \R {\mbox{\rm \hbox{I\kern-.15em\hbox{R}}}}

\definecolor{amethyst}{rgb}{0.6, 0.4, 0.8}
\definecolor{darkpastelgreen}{rgb}{0.01, 0.75, 0.24}

\begin{document}

\title{Towards automatic digital garment assembly}

\author{Alejandro Rodríguez~\orcidlink{0000-0001-8085-1523}
\thanks{Alejandro Rodríguez is with SEDDI. E-mail: alejandrora88@gmail.com.}
}
\maketitle

\begin{figure}
    \centering
    \includegraphics[width=1\linewidth]{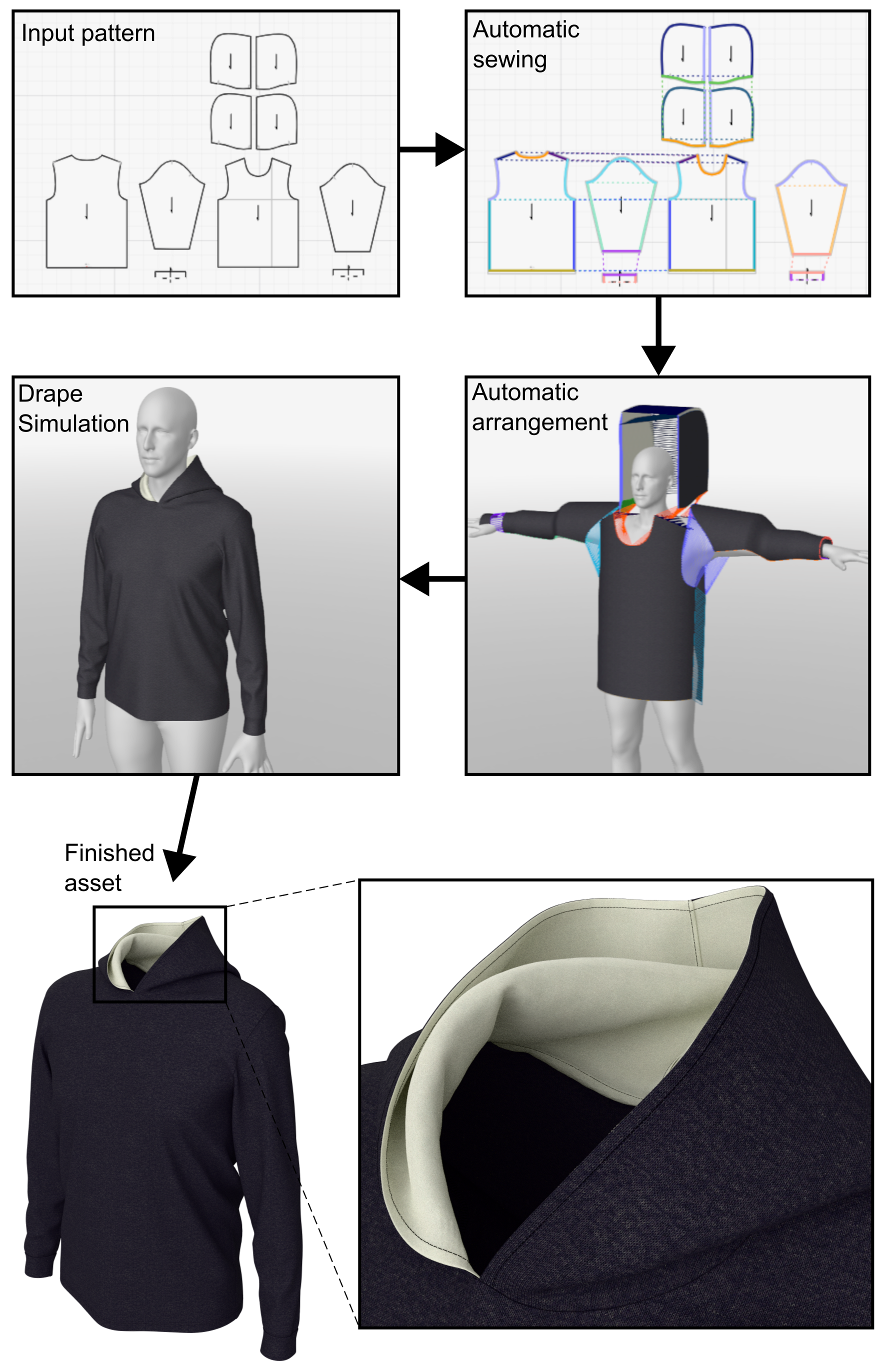}
    \caption{Application of our proposed solutions integrated into an apparel CAD system. The input 2D pattern, shown at the top left, is processed by our automatic sewing and arrangement models to assemble the garment into a simulation-ready configuration. After draping, the True Seams process \cite{rodriguez2022true} creates the final, detailed geometry. The entire process is automatic, going from 2D pattern to finished 3D asset without manual adjustments required.}
    \label{fig:teaser}
\end{figure}

\begin{abstract}
Creating digital garments from existing two-dimensional patterns remains a time-consuming process. Although patterns are widely available in digital form, their assembly in garment CAD software involves manual definition of sewing operations and arrangement of the pattern panels for drape simulation. We present two learning-based approaches that automate both tasks using the 2D vector pattern as their sole input. Our automatic sewing model operates at the segment level, leveraging a Transformer encoder with multiple prediction heads. Our novel architecture supports many-to-many, ordered segment correspondences, as well as hems, overlapping seams needed for multilayer garments, and construction properties such as seam allowances and stitch types, required for the finished virtual garment but neglected by previous methods. A complementary panel-level model predicts body-region assignments, layers, and placement coordinates to initialize the drape simulation. We evaluate both models on a curated dataset of 4,035 professionally assembled garment instances. The sewing model exactly recovers all seam and hem topologies of 63.3\% of test garments, while the arrangement model meets the specified placement criteria for all panels in 66.2\% of test garments. We also integrate both methods into garment CAD software, demonstrating automatic assembly of successfully predicted garments.
\end{abstract}


\section{Introduction}
\label{sec:introduction}

\IEEEPARstart{T}he apparel industry increasingly relies on digital workflows for garment design and production. Two-dimensional garment patterns are commonly developed and maintained in CAD systems that support pattern editing, size grading, and integration with automated cutting systems.

Digital garments also serve as 3D assets for gaming, advertising, and filmmaking, while enabling virtual prototyping in the apparel industry to accelerate design iteration and reduce reliance on physical samples. However, assembling a finished 3D garment from an existing 2D pattern still requires substantial manual work from expert digital fashion designers using apparel CAD software such as CLO \cite{Clo}, Browzwear \cite{Browzwear}, or Style3D \cite{Style3D}. Designers must first define all the sewing operations and their construction properties, such as stitch types and seam allowances, to reproduce the details of the physical garment. They then arrange the pattern panels around a target body and run a physical simulation to obtain the final drape.
	
This manual process has become a real bottleneck for digitization efforts that the industry faces, prompting extensive research aimed to its automation. Many proposed learning-based solutions rely solely on the 2D pattern as inpnt, demonstrating that these patterns encode valuable information for inferring sewing configurations and 3D properties. Nevertheless, existing solutions still have inherent limitations and tackle only a subset of garments, preventing their application to many real patterns that are very common in apparel.
	
In this work, we propose a novel method for automatic sewing of 2D vector patterns, leveraging deep learning techniques to design and train a custom model that addresses some of the limitations found in previous work. Our proposed architecture targets a broader set of more complex garments by supporting ordered, multi-segment sewing operations including overlapping seams, required for multilayer garments, and hems, typically applied to the free edges of a garment. The same architecture also enables the inference of seam properties along with the seams themselves, neglected by previous works that only address the topology of the seams, but required nonetheless for the detailed generation of finished assets.
	
We also propose a separate architecture for automatic panel identification and arrangement to initialize the drape simulation from a favorable configuration.
	
Finally, we show how these two methods can be integrated with existing workflows and how they can significantly reduce the amount of manual assembly input required, in many cases even enabling full automation from the 2D pattern to the final 3D asset.
\section{Related Work}
\label{sec:related-work}

Recent research has addressed several stages of digital garment creation, including pattern generation, sewing prediction, and initialization for physical simulation. Wang et al.~\cite{wang2018training} learn a shared shape space linking sewing patterns and draped garments, enabling sketch-guided design. NeuralTailor~\cite{korosteleva2022neuraltailor} reconstructs sewing patterns and their seams from garment point clouds, while Liu et al.~\cite{liu2023towards} and PanelFormer~\cite{chen2024panelformer} recover patterns from assembled garment images. More recently, Cao et al.~\cite{cao2026learning} combine an implicit panel representation with latent flow matching to support pattern generation, image-conditioned reconstruction, completion, and refitting. These methods address the recovery or generation of pattern geometry, and many of them include seam recovery solutions \cite{korosteleva2022neuraltailor,liu2023towards,chen2024panelformer,cao2026learning}, but they take advantage of some form of 3D assembly information present in the input. Our work focuses on assembling an existing 2D vector pattern, using the pattern itself as sole input, retaining its geometry and predicting the sewing operations and arrangement required for subsequent simulation.

Inferring sewing relationships from existing patterns has a longer history. In their seminal work, Berthouzoz et al.~\cite{berthouzoz2013parsing} combine machine learning and integer programming to parse annotated sewing patterns and infer how their panels are sewn together. Their method relies on semantic information such as panel labels and notches to resolve assembly ambiguities, but demonstrated that effective seam inference was possible without 3D assembly input information. AutoSew~\cite{rios2026autosew} instead predicts seam relationships from geometric input without requiring semantic annotations. It combines a graph neural network with differentiable optimal transport to infer partial, multi-edge seam assignments, addressing the limitations of one-to-one segment correspondences found in previous works. This capability is important for production patterns, as even the most basic garments exhibit multi-segment seams.

Wang et al.~\cite{wang2026learning} also address sewing prediction from 2D panel geometry through a two-stage graph-based framework, using image-based representations of the panels. They first infer panel semantics and connectivity, then reconstruct detailed boundary correspondences using local geometric features and garment context. Their method also supports many-to-many correspondences.

GarmageNet~\cite{li2025garmagenet} addresses garment generation and assembly within a broader multimodal generative framework that uses a shared 2D-3D image representation for garment panels. It supports generation conditioned on diverse inputs, including an actual 2D pattern, but the final garment geometry is reconstructed from its underlying rasterized representation and does not guarantee exact preservation of the input vector geometry. Our method instead operates directly on the supplied Bézier segments, predicting sewing operations while preserving the original pattern geometry.

Beyond boundary correspondences, detailed garment assembly requires additional sewing operations and construction properties. As discussed in \S~\ref{sec:garment-patterns-and-sewing-operations}, facing seams can overlap with regular seams on the same boundary segments, while hems finish boundaries without joining separate panels. We address these operations within a unified prediction model that recovers ordered, multi-segment regular and facing seams, and hems, together with properties such as seam allowances, folding directions, and stitch types. These outputs support both garment assembly and the generation of seam and finishing details.

A suitable initial arrangement is also necessary for a reliable drape simulation. Liu et al.~\cite{liu2024automatic} combine learned classification, heuristics, and numerical optimization to provide a fine-grained arrangement of the garment panels around a target body, taking the sewing relationships as input. GarmageNet~\cite{li2025garmagenet} likewise provides detailed 3D initialization through its generated geometry. We predict arrangement directly from the 2D pattern using a novel model architecture. Its outputs specify body-region assignments, layers, and placement coordinates in the arrangement space provided by the CAD system, allowing predictions to be inspected and corrected through the existing interface.
\section{Garment Patterns and Sewing Operations}
\label{sec:garment-patterns-and-sewing-operations}

A 2D garment pattern defines a set of panels, each described as a vector outline, typically a piecewise Bézier curve \cite{farin2002curves} comprising segments of degree one to three. Seam operations are conceptually simple, as they join boundary segments of the different panels together, but complexity increases quickly even in the most basic garments.
	
The simplest seam joins two segments one-to-one. This very common seam is addressed by early works in the field \cite{liu2023towards,chen2024panelformer}, but equally common are one-to-many, or many-to-many segment-joining seams, as Ríos-Navarro et al. correctly identified \cite{rios2026autosew}. Earlier works use synthetic datasets that avoid these more complex seams by geometrically splitting segments and panels to always accommodate one-to-one correspondences \cite{korosteleva2021generating,GarmentCode2023,GarmentCodeData:2024}. Unfortunately, that is not an option when working with industry-grade patterns, as the seams are not known beforehand. Moreover, it is actually not uncommon to find 2D patterns with segments split in two or more parts, either intentionally by the pattern maker to add markings for different reasons (adjustment points, logo placements, etc.) or unintentionally due to the tools used to create the pattern, including export or conversion tools that can sometimes split segments based on geometric rules. Adding robustness against such occurrences is paramount for any solution that applies to industry-grade patterns.

\begin{figure}
    \centering
    \includegraphics[width=\linewidth]{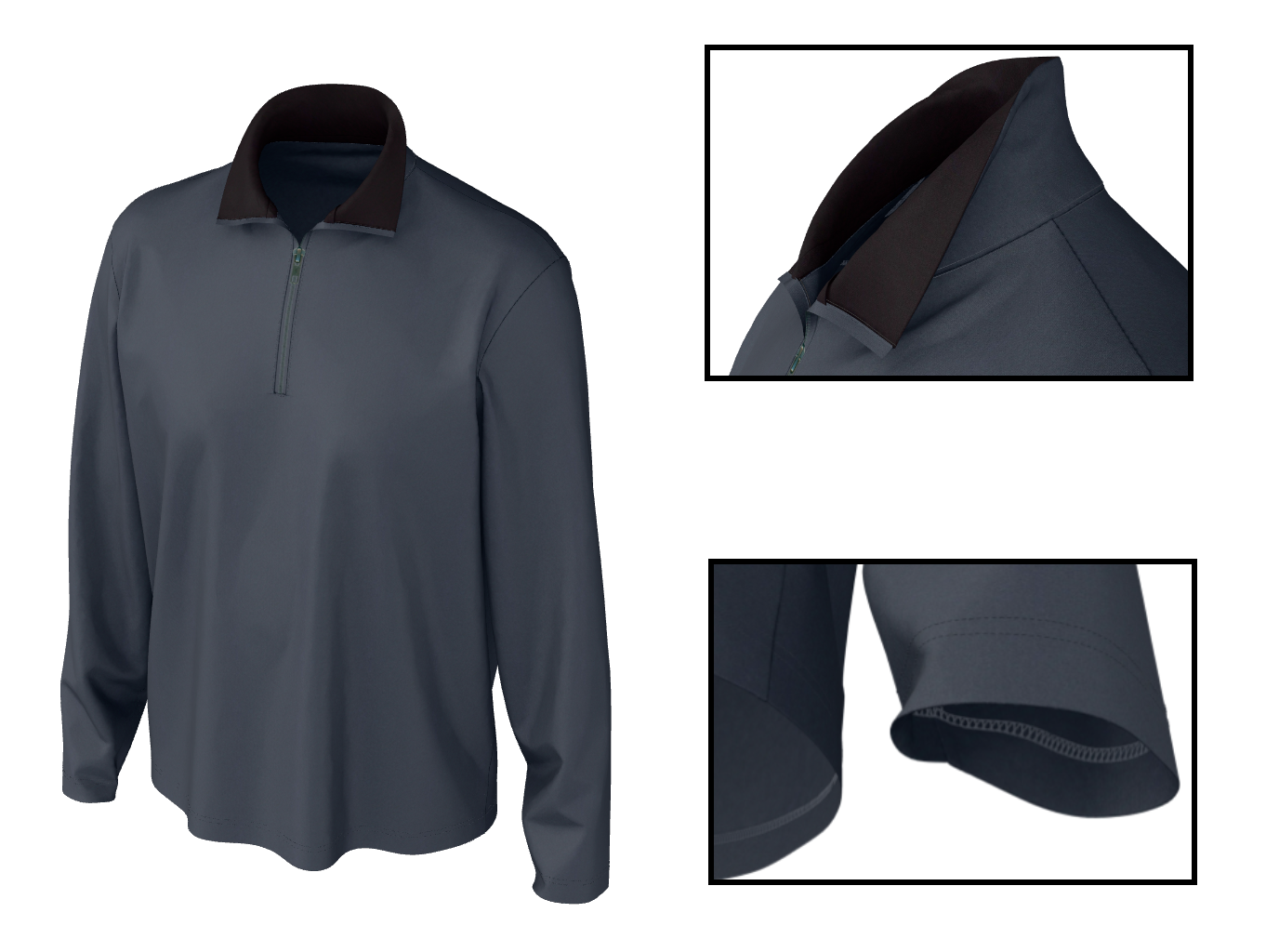}
    \caption{Example of a finished garment with hems and facing seams. The collar has two layers of fabric joined together by facing seams before being sewn to the torso panels with a regular seam. The sleeves and the shirt bottom are finished with hems.}
    \label{fig:facing-hem-examples}
\end{figure}

All of the seams described above can be cataloged as \emph{regular} seams, and have been the main target of previous works, but they are not the only sewing operation commonly found in garments. Another very common sewing operation joins two panels one on top of the other (instead of one alongside the other) to create layered garments. We refer to these seams as \emph{facing} seams. A segment involved in a facing seam can still be involved in a subsequent regular seam, as shown in Figure \ref{fig:facing-hem-examples}. More examples are displayed in Figure \ref{fig:dataset-samples}.

Some of the boundary segments of the panels are not sewn to any other, but that does not necessarily mean that they simply remain untouched. They are typically finished with another sewing operation known as a \emph{hem}. An example is also shown in Figure \ref{fig:facing-hem-examples}. These are not required to obtain simulation-ready garments, since they do not modify the topology of the panels, but they become relevant when the goal is to produce a finished asset, as they impact the final look of the garment.

Our proposed solution predicts regular seams, as well as facing seams and hems, largely ignored by previous solutions.
\section{Automatic Sewing}

\begin{figure*}[h!]
    \centering
    \includegraphics[width=\linewidth]{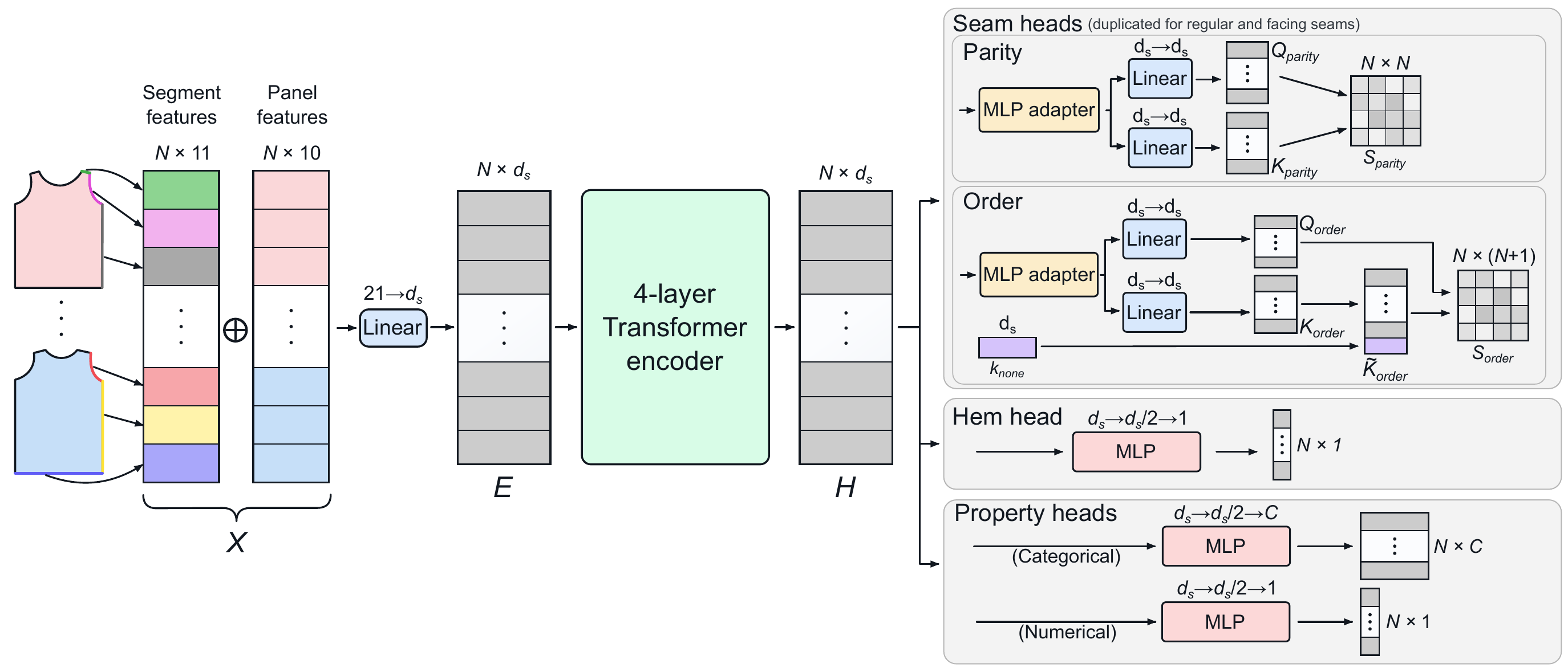}
    
    \caption{Proposed automatic sewing model architecture. Per-segment features $X$, including panel-level features, are input to the model. They are projected into embeddings $E$ and refined by a Transformer encoder with full self-attention. Specialized heads then process the enriched embeddings $H$ to reconstruct the topology and properties of the different sewing operations.}

    \label{fig:auto-sewing-model}
\end{figure*}

\begin{figure}
    \centering
    \includegraphics[width=0.7\linewidth]{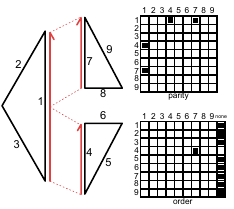}
    \caption{Toy pattern example with a single multi-segment seam, showing the corresponding target parity and order matrices encoding the seam topology.}
    \label{fig:seam-targets}
\end{figure}

We approach automatic sewing as a segment-level learning problem. We propose a novel architecture that takes all garment pattern segments as input through a backbone encoder that computes contextually rich segment embeddings. These embeddings are then fed to multiple prediction heads with specialized configurations for each of the downstream tasks. A diagram of the full architecture is shown in Figure \ref{fig:auto-sewing-model}.
	 
For proper contextual enrichment, local and global geometric information must be shared among the segments. The Transformer architecture typically requires positional embeddings to reason about relative positions among input tokens \cite{vaswani2017attention}. Ríos-Navarro et al. \cite{rios2026autosew} identify this problem and resort to using a graph neural network using intra-panel topology neighborhood to create the edges of the graph. This creates a significant inductive bias that enriches segment embeddings with intra-panel information, but it impedes inter-panel contextual enrichment. We make the observation that the basic segment features, i.e., its control points, implicitly encode enough information to deduce topological connections, as the last control point of a segment is identical to the first control point of its following neighbor by construction. Moreover, we also note that in many patterns, it would be useful to share context with segments from different panels. In particular, multilayer garments typically contain duplicated panels that could lead to very similar embeddings, as some of the examples in Figure \ref{fig:result-samples}. In those cases, awareness of the surrounding panels and their relative positions can further enrich the segment embeddings and help to disambiguate. These observations, together with the demonstrated effectiveness and versatility of Transformer architectures across diverse tasks and modalities \cite{dosovitskiy2021an,lu2022frozen} motivate us to leverage it for our architecture. As we show in \S \ref{sec:results}, the experiments support this design choice.

\subsection{Segment input features}
We first compute a set of per-panel features that include the center of the panel, its area, the length of its perimeter, and its centered axis-aligned bounding box minimum, maximum, and size, for a total of 10 features. Then, we compute a set of per-segment features: a one-hot encoding of the degree of the Bézier segment, and the panel-centered 2D coordinates of up to four control points (two points in the case of a linear Bézier, four points in the case of a cubic Bézier), setting to zero the ones not used depending on the degree, for a total of 11 features. Finally, for each segment, we concatenate its segment features with its corresponding panel features for a final feature vector of 21 values, implicitly grouping same-panel segments by their shared panel features. All vectors are stacked into an input segment tensor $X \in \mathbb{R}^{N \times 21}$.

\subsection{Encoder}
The backbone of our model is a Transformer encoder that processes all $N$ segments at once. The input feature vectors are first projected up to an embedding tensor $E \in \mathbb{R}^{N \times d_s}$ through a learnable linear layer. These embeddings are then passed through a standard four-layer Transformer encoder with four attention heads per layer, ReLU activations, feed forward dimension = $4d_s$, pre-normalization \cite{xiong2020layer} and dropout = 0.1. Based on our earlier observations, the topology information is implicitly contained in the segment feature vectors, and panel-awareness is introduced by the concatenated panel features. We therefore simply enable full self-attention without masking or additional positional encodings, letting the model learn the optimal attention patterns for the task and output the refined contextual segment embeddings $H \in \mathbb{R}^{N \times d_s}$. We set $d_s = 512$.

\subsection{Seam prediction heads}
Focusing first on regular seams, we are addressing the general problem of many-to-many segment correspondences: a seam is composed of two sides, each side with an ordered list of segments. We formulate it as a pairwise segment similarity computation problem, but, in contrast to prior work, we compute two separate pairwise segment similarity scores. We refer to the first one as \emph{parity} score and it computes logits for whether a segment pair belongs to the opposing sides of a seam, predicting then the sets of segments for each seam side for all seams. The second one, referred to as \emph{order} score, computes logits for whether a segment goes immediately before another one on their shared seam side, predicting then the order of segments on every seam side. An illustrative example is shown in Figure \ref{fig:seam-targets}.

These two computations actually require different similarities among the same segments. The embeddings produced by the encoder pass through separate task-specific MLP adapters, each consisting of a LayerNorm followed by two fully connected $d_s$-dimensional layers with GELU activations \cite{hendrycks2016gaussian}. These task-specialized embeddings are then linearly projected into query and key matrices, $Q$ and $K$.

\emph{Parity} scores are commutative, i.e., score for segment pairs ${(s_i,s_j)}$ and ${(s_j,s_i)}$ should be the same, so we compute the symmetric score matrix $S_{parity} \in \mathbb{R}^{N \times N}$ as

\begin{equation}
    S_{parity} = \frac{(Q_{parity}K_{parity}^T + K_{parity}Q_{parity}^T)}{2\sqrt{d_s}}.
\end{equation}

For the \emph{order} score, each segment must select either a single segment or none. We enable the latter option by appending a learned key vector to the K matrix

\begin{equation}
\tilde{K}_{order} = 
\begin{bmatrix}
K_{order} \\
k_{none}
\end{bmatrix}
\end{equation}

and computing the score matrix $S_{order} \in \mathbb{R}^{N \times (N+1)}$ as

\begin{equation}
S_{order} = \frac{Q_{order}\tilde{K}_{order}^T}{\sqrt{d_s}}.
\end{equation}

During training, we exclude the lower-triangular and diagonal entries of $S_{parity}$ from the pairing loss. During inference, those same entries are excluded from the pairing predictions. In the case of $S_{order}$, the diagonal entries are set to $-\infty$ during both training and inference, preventing a segment from being selected as its own successor.

Facing seams can involve the same segments as regular seams. Since they present the same many-to-many segment correspondence problem, we simply duplicate the two proposed heads with their own adapters to accommodate the new tasks.
    
\subsection{Hem prediction head}
Hems do not join two sides, but a single side of multiple ordered segments. For this goal, we add a binary classification head, a 2-layer MLP that first projects the input embedding to $d_s / 2$, applies a ReLU activation, and then maps the resulting representation to logits. This head  processes $H$ row-wise to output whether or not a segment belongs to a hem. For the order, we could add another \emph{order} score head, but, since hems and regular seams can never overlap, we can actually reuse the \emph{order} score head used for regular seams, saving some memory and computation.

\subsection{Seam property heads}
In addition to the involved segments, sewing operations also require specifying additional properties that affect the final assembly and look of the garment, such as seam allowance width, folding direction, seam finishing type, stitch type, etc. We add 2-layer MLP classification heads to predict categorical properties and 2-layer MLP regression heads to predict numerical properties, directly acting row-wise on $H$. Since we estimate these properties per segment, we broadcast all target properties of a seam to each segment comprising that seam for training. Segments that do not receive a particular target property are excluded from the loss computation for that property.

\subsection{Loss}
\label{sec:auto-sewing-loss}

The proposed architecture is fully differentiable, and we train it end to end.

The $\mathcal{L}_{\mathrm{parity}}$ loss used for each \emph{parity} score matrix is computed with standard binary cross-entropy loss over the matrix logits. 
		
The $\mathcal{L}_{\mathrm{order}}$ loss for each \emph{order} score matrix is a standard cross-entropy loss applied row-wise.

Hem membership uses binary cross-entropy,
denoted by $\mathcal{L}_{\mathrm{hem}}$, while hem ordering is already
supervised through the shared regular-seam order head.

Categorical and numerical seam properties use standard cross-entropy and mean squared error, respectively.
We denote the sum of all individual property losses by
$\mathcal{L}_{\mathrm{prop}}$.

All the losses are added into a total loss:

\begin{equation}
\begin{aligned}
\mathcal{L}_{\mathrm{total}}
={}&
\sum_{t\in\{\mathrm{regular},\mathrm{facing}\}}
\left(
\mathcal{L}_{\mathrm{parity}}^{(t)}
+\mathcal{L}_{\mathrm{order}}^{(t)}
\right)\\
&+\mathcal{L}_{\mathrm{hem}}
+0.01\,\mathcal{L}_{\mathrm{prop}}.
\end{aligned}
\end{equation}

We assign reconstruction losses 100 times the weight of each
property loss, prioritizing sewing operation reconstruction, as it is the more critical and sensitive target.

\subsection{Seam reconstruction}
Reconstructing sewing operations from the model output is a straightforward process.

For regular and facing seams, we first create a graph with all segments as nodes and positive pairs (i.e., pairs with positive logits) in the corresponding \emph{parity} matrix as edges. We detect connected components using a standard Union-Find algorithm \cite{tarjan1975efficiency}. For each connected component, we then use graph bipartition to detect the two sides of the seam. Finally, we order the segments of each seam side using the top row-wise entries in the corresponding \emph{order} matrix. If a connected component is not bipartite, or if it is bipartite but either side has a faulty ordering (cycles, competing predecessors, or disconnected chains), we completely discard it. 

Hems are reconstructed in a similar fashion, creating a graph with only the subset of segments flagged as hems and the top row-wise entries in the \emph{order} matrix as edges. We then compute connected components, each being a final hem, and order each one using the same \emph{order} matrix, similarly discarding components with faulty ordering.

Once each sewing operation is reconstructed, its properties are collectively determined by its constituent segments through majority voting for categorical properties and averaging for numerical properties.
\section{Automatic Arrangement}

\begin{figure*}[h!]
    \centering
    \includegraphics[width=\linewidth]{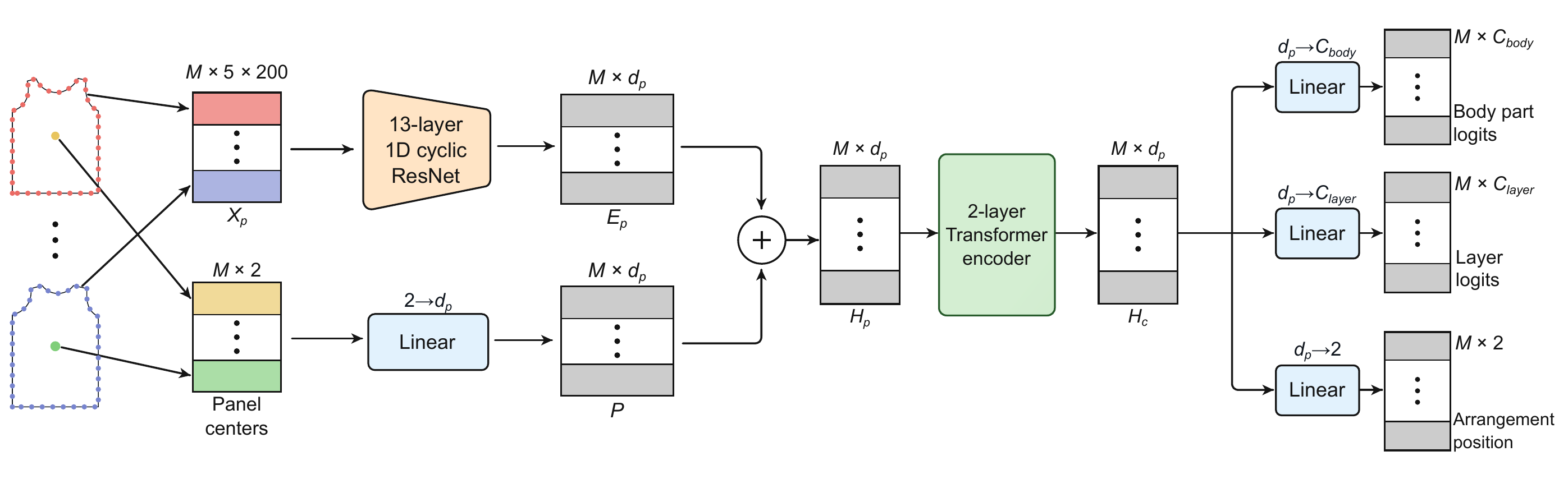}
    
    \caption{Proposed architecture for the automatic arrangement model. The sampled boundaries of the target garment panels are processed by a 1D cyclic ResNet to produce per-panel embeddings $E_p$. Positional embeddings $P$ are computed and added to enrich the panel embeddings $H_p$. They get further refined with contextual awareness by a Transformer encoder with full self-attention. The final embeddings $H_c$ are then passed through simple linear projections to generate the per-panel arrangement configurations.}

    \label{fig:auto-arrangement-model}
\end{figure*}

\begin{figure}
    \centering
    \includegraphics[width=0.9\linewidth]{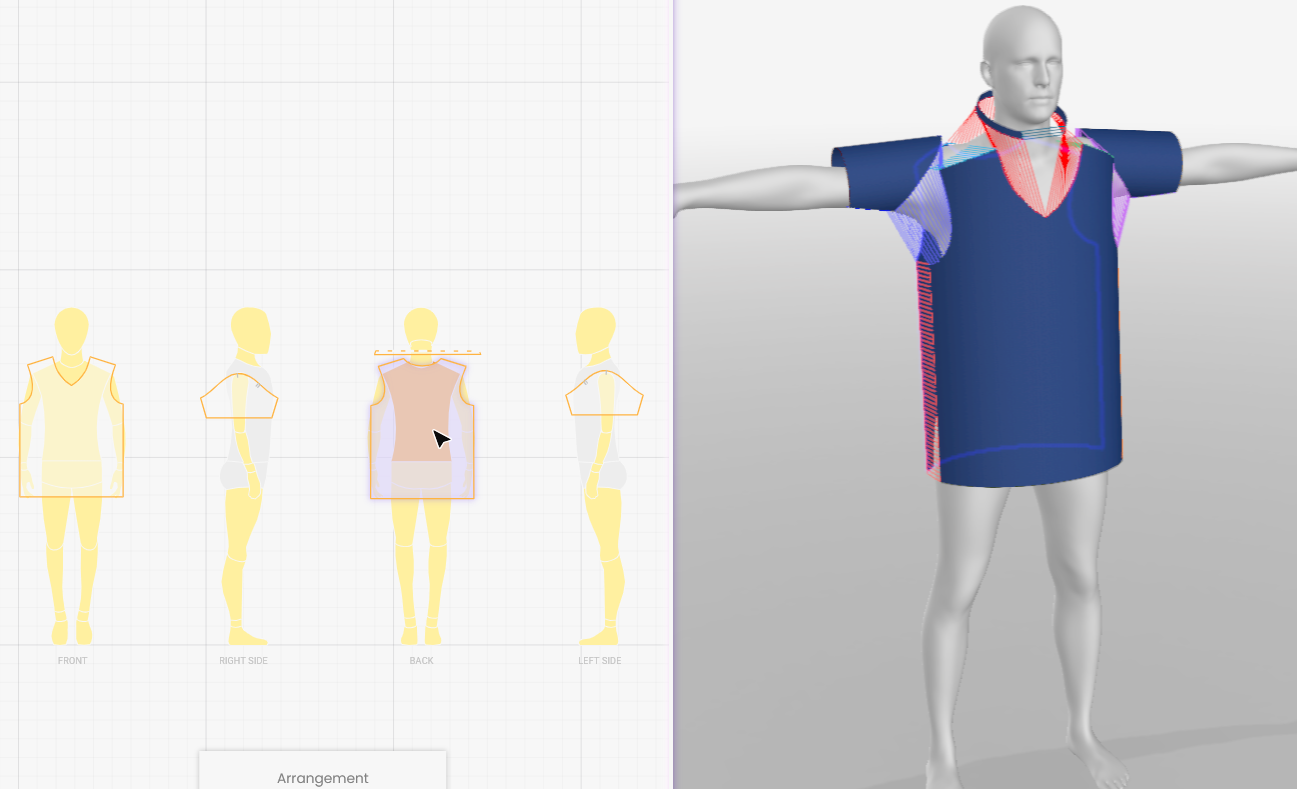}
    \caption{Arrangement tool provided by the CAD software. On the left side, a 2D space allows users to assign the different pattern panels to body regions. These assignments map the panel geometries to the 3D space on the right, wrapping them around the body, properly arranging them for a successful drape simulation.}
    \label{fig:arrangement-tool}
\end{figure}

The garment arrangement to initialize the drape simulation is typically done by placing the panels in 3D around the target body. Instead of manually placing the panels in the 3D space, CAD applications usually provide an easier parametric space that maps to the 3D space. In our particular CAD software, each panel is assigned to one of a set of predefined body regions, assigned to one layer, and given a position in a 2D \emph{arrangement} space as shown in Figure \ref{fig:arrangement-tool}, mapping them to a corresponding 3D configuration based on the pose and shape of the target body. While easier than a raw 3D arrangement, it is still manual.
	
We propose a different architecture to automate this assignment.

\subsection{Model architecture}
This problem is more naturally addressed at the panel level instead of the segment level, so we can resort to a smaller panel-wise encoder to capture the relevant panel information.

We propose an architecture for panel-level prediction that combines a one-dimensional cyclic residual network with a Transformer encoder that processes all garment panels at once. Figure \ref{fig:auto-arrangement-model} shows a diagram of the model.

We prepare the model input by discretizing each of the $M$ garment panels as a multichannel 1D signal, uniformly sampling its boundary. Each sample consists of panel-centered position, normal, and local curvature, and we take 200 samples. All garment panel signals are stacked into the input tensor $X_p \in \mathbb{R}^{M \times 5 \times 200}$.

The model first generates $E_p \in \mathbb{R}^{M \times d_p}$ panel embeddings by processing each panel signal with a 13-layer 1D ResNet. Our ResNet follows the design principles of conventional image-based ResNets \cite{he2016deep}, adapted to the one-dimensional and cyclic nature of our garment outlines: 1D instead of 2D convolutions and \emph{circular} padding for all convolution operations to account for the circular connections between the first and last sample of the signal. We set $d_p = 256$.

Each independently computed panel embedding is then enriched with positional information. We linearly project the center of each panel into $p_i \in \mathbb{R}^{d_p}$, $i \in \{1,...,M\}$, stack them together and add these projections to the panel embeddings $H_p = E_p+P$. $H_p$ is then processed by a two-layer Transformer encoder with full self-attention, four attention heads per layer, ReLU activations, feed forward dimension = $d_p$, pre-normalization and dropout = 0.1, producing $H_{c}$ by further refining the representations of the panels through contextual awareness.

Finally, three independent linear prediction heads process the final embeddings $H_{c}$ to output the categorical body region assignment logits, the layer assignment logits, and the 2D \emph{arrangement} position.

The model is trained end to end using a total loss given by the sum of the cross-entropy losses for the body region and layer assignments and the mean squared error (MSE) loss for the 2D \emph{arrangement} position:
\begin{equation}
    \mathcal{L}
    =
    \mathcal{L}_{\mathrm{CE}}^{\mathrm{body}}
    +
    \mathcal{L}_{\mathrm{CE}}^{\mathrm{layer}}
    +
    \mathcal{L}_{\mathrm{MSE}}.
\end{equation}

For each multiclass assignment, the cross-entropy loss is defined as
\begin{equation}
    \mathcal{L}_{\mathrm{CE}}
    =
    -\frac{1}{M}
    \sum_{i=1}^{M}
    \sum_{c=1}^{C}
    y_{ic}
    \log\left(\hat{y}_{ic}\right),
\end{equation}
where $y_{ic}$ is the ground-truth one-hot label and $\hat{y}_{ic}$ is the predicted probability for class $c$. The MSE loss for the predicted 2D position is given by
\begin{equation}
    \mathcal{L}_{\mathrm{MSE}}
    =
    \frac{1}{M}
    \sum_{i=1}^{M}
    \left\|
        \mathbf{p}_i - \hat{\mathbf{p}}_i
    \right\|_2^2,
\end{equation}
where $\mathbf{p}_i, \hat{\mathbf{p}}_i \in \mathbb{R}^2$ denote the ground-truth and predicted \emph{arrangement} positions, respectively.

\section{Dataset}
\label{sec:dataset}
We train our models on a curated dataset of virtual replicas of real garments created by a team of professional pattern makers. All garments have been manually assembled by creating the proper sewing operations, including seam properties corresponding to the real counterparts, and they have been manually arranged and draped on corresponding virtual human models.

\begin{figure*}[h!]
    \centering
    \includegraphics[width=\linewidth]{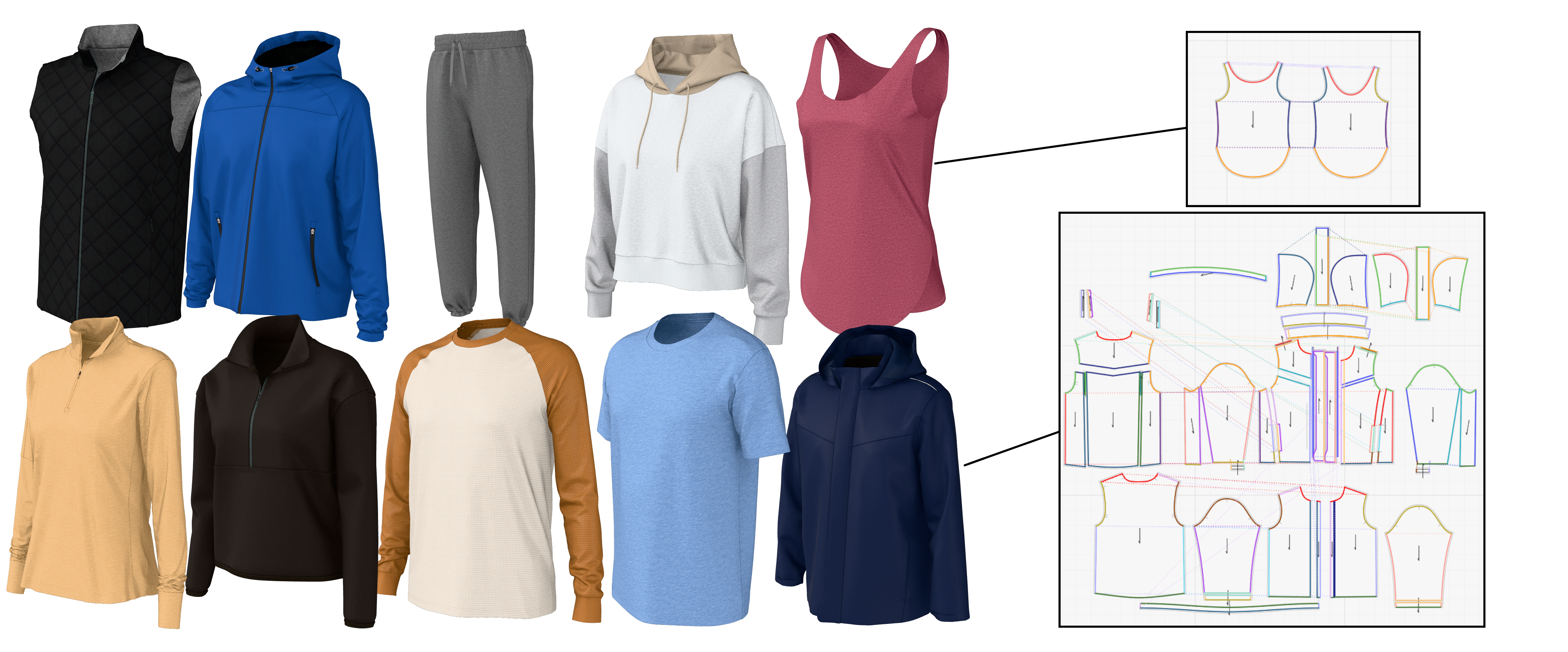}
    
    \caption{Examples of garments of different complexity in our dataset. The pink tank, shown at the top right, has a very simple pattern with 2 panels, 4 regular seams and 4 hems. The navy jacket, shown at the bottom right, has a more complex pattern with 47 panels, 45 regular seams and 23 facing seams.}

    \label{fig:dataset-samples}
\end{figure*}

The dataset consists of 4,035 garments representing 1,089 distinct styles, across various categories, such as t-shirts, tanks, sweatshirts, polo shirts, sweatpants, shorts, vests, jackets, etc. Of these styles, 516 have multiple size variations, ranging from 2 to 10 sizes per style, while 573 are single size. Table \ref{table:dataset-content} summarizes the contents of the dataset, and Figure \ref{fig:dataset-samples} shows some examples of garments and corresponding patterns.

\begin{table}
    \centering
    
    \caption{Dataset statistics on garment panels and sewing operations. }
    
    \begin{tabular}{lrrrr}
    \hline
                    &   Average &   Min &   Max &   Total \\
    \hline
     Garments       &  - &  - & - &      4,035  \\
     Panels         & 8.68 &  2 & 52 &     35,038  \\
     Regular seams  & 13.19 &  4 & 49 &      53,220  \\
     Facing seams       & 1.56 &  0 & 25 &     6,312  \\
     Hems           & 2.54 &  0 & 14 &      10,243  \\
    \hline
    \end{tabular}

    \label{table:dataset-content}
\end{table}

We use standard splits of 80\% for training, 10\% for validation, and 10\% for testing. Since our dataset contains full size ranges for many of the styles, we ensure that the entire size range falls on the same split, as different sizes of the same style typically only differ in some non-uniform scaling of the panels and could pollute the evaluation metrics.

\subsection{Data augmentation}
We augment the dataset using three randomized online augmentation strategies to improve generalization.

\textbf{Starting segment -} We apply a cyclic rotation with a random offset to the segments of each panel, updating the corresponding sewing operations, to prevent biasing our models towards specific pattern panel definition tendencies. 

\textbf{Segment splitting -} We randomly split panel segments into two, also updating the sewing operations accordingly, to emulate the splittings found in real patterns, as discussed in \S\ref{sec:garment-patterns-and-sewing-operations}. We split segments with a $10\%$ probability at a random parametric location in the $[0.3,0.7]$ range to avoid ill-defined segments.

\textbf{Uniform scaling -} We scale the whole garment uniformly by a random factor in the $[0.9,1.1]$ range to add size variability.
\section{Results}
\label{sec:results}

We have implemented and trained both proposed models in PyTorch using the AdamW optimizer \cite{loshchilov2017decoupled} with 0.01 weight decay on an NVIDIA A10 GPU. For both models, we use a batch size of 16 during training. For the automatic sewing model, we use a learning rate of $5 \times 10^{-5}$. Training runs at 11.7 seconds per epoch and its validation performance plateaus after $\sim$900 epochs. For the automatic arrangement model we use a learning rate of $5 \times 10^{-4}$, training runs at 4.7 seconds per epoch and validation performance plateaus after $\sim$180 epochs.
	
Inference is very fast, with each model running in less than 60 ms on the largest input pattern in our test set, including the sewing operations reconstruction procedure in the case of the automatic sewing model, enabling easy integration in interactive workflows.

In this section, we provide a quantitative evaluation of the performance of both models. We also show additional ablation experiments that support some of our decisions and findings. Finally, we integrate our models into a digital garment creation CAD system and show how we can go from 2D pattern to 3D assembled garments and, in some cases, finished assets in a fully automatic manner.

\subsection{Automatic sewing evaluation}

To our knowledge, previous sewing prediction methods do not jointly support regular seams, overlapping facing seams, and hems, preventing a direct comparison of recovery performance over the complete set of sewing operations considered here. For quantitative evaluation of the automatic sewing model performance, we compute the percentage of sewing operations recovered exactly (considering only topology, i.e., correct per-side ordered segments, ignoring seam properties) over the total in the test set, as well as perfectly reconstructed garments, meaning all their regular seams, hems and facing seams are recovered and no extra sewing operations are created. We find a perfect recovery of 63.28\% of the garments. Looking into the imperfectly recovered garments that would require manual intervention to add or remove sewing operations, we find that 73.3\% of them contain 4 or fewer bad sewing operations, and only 7.6\% of them contain 10 or more bad sewing operations, so even when the full garment is not recovered, most of it typically is, saving the user a significant portion of manual work. Some examples are shown in Figure \ref{fig:result-samples}.

\begin{table}
    \centering
    
    \caption{Results of the automatic sewing experiments on the test set, with perfect recovery percentages for whole garments and for regular seams, facing seams and hems. Corresponding results for the ablation experiments are also reported.}
    
    \begin{tabular}{lrrrr}
        \hline
        & \multicolumn{4}{c}{Perfect recovery (\%)} \\
        \cline{2-5}
        Method & Garments & R. seams & F. seams & Hems \\
        \hline
        Ours                & \textbf{63.28} & \textbf{91.52} & 77.68 & 86.12 \\
        Neighbor att.     & 48.31 & 87.06 & 71.10 & 83.18 \\
        Extra Bézier feat.  & 58.45 & 90.72 & 75.61 & \textbf{86.73} \\
        Negative mining        & 60.63 & 91.46 & \textbf{78.29} & 85.36 \\
        \hline
    \end{tabular}
     \label{table:auto-sewing-results}
\end{table}

Regarding each type of sewing operation, we find a perfect recovery of 91.52\% for regular seams, 77.68\% for facing seams and 86.12\% for hems. A summary of the results is shown in Table \ref{table:auto-sewing-results}. Interestingly, regular seams perform the best, for which our dataset contains the most examples, and facing seams perform the worst, with the lowest number of examples in our dataset. These results align with the empirical correlation found between the size of the dataset and the performance of the model, now prevalent in deep learning \cite{hestness2017deep}, suggesting that a larger dataset with more examples of underrepresented sewing operations would improve model performance.

\begin{table}
    \centering
    
    \caption{Metrics corresponding to the evaluation outputs of the different automatic sewing model heads. We report accuracy (acc.), precision (prec.), recall and F1 for regular (R.), facing (F.) and hem classification tasks. We also report mean absolute error (MAE) for seam allowance regression. Properties are evaluated per labeled segment.}
    
    \begin{tabular}{lrrrrr}
    \hline
    Task & Acc. & Prec. & Recall & F1 & MAE (cm) \\
    \hline
    R. $S_{parity}$ & 99.79 & 95.74 & 95.97 & 95.85 & - \\
    R. $S_{order}$  & 96.63 & 96.63 & 96.63 & 96.63 & - \\
    F. $S_{parity}$ & 99.93 & 93.12 & 90.93 & 92.01 & - \\
    F. $S_{order}$  & 99.15 & 99.15 & 99.15 & 99.15 & - \\
    Hem segment    & 99.13 & 96.10 & 90.14 & 93.03 & - \\
    Fold side      & 85.94 & 85.93 & 85.94 & 85.93 & - \\
    R. type        & 85.85 & 85.35 & 85.85 & 85.48 & - \\
    R. stitch type & 83.23 & 83.20 & 83.23 & 82.98 & - \\
    Hem type       & 94.27 & 93.18 & 94.27 & 93.28 & - \\
    F. type        & 86.69 & 87.77 & 86.69 & 86.86 & - \\
    F. stitch type & 92.09 & 91.70 & 92.09 & 91.51 & - \\
    Allowance      & - & - & - & - & 0.15 \\
    \hline
\end{tabular}

    \label{table:property-metrics}
\end{table}

Table \ref{table:property-metrics} reports additional metrics for the different model heads, also including seam properties for a more detailed breakdown. Note that the F1 score for some of the properties is notably lower. We believe that this is due to the stylistic nature of some of these properties: some construction choices such as the stitching applied to regular seams, particularly in visible areas of the garment, have a stylistic component that varies depending on brand or season among other variables, making them harder to derive solely from the pattern. Other construction elements, on the other hand, are typically not visible, such as the stitching applied to facing seams, so their choice responds more to structural concerns and can be more easily derived from the pattern alone.
    
\subsubsection{Topology-constrained attention}

We compare our choice of full self-attention with a model using masked attention to restrict segments to only attend to direct neighbors in their panel, loosely replicating the inductive bias exhibited by the architecture proposed by Ríos-Navarro et al.\cite{rios2026autosew}.

Table \ref{table:auto-sewing-results} summarizes the results, showing a noticeable drop in reconstructed sewing operations as well as full garments. These results support our hypotheses that the raw Bézier features contain sufficient information for the Transformer model to infer relevant topological relationships and that enabling full inter-panel and intra-panel attention improves overall performance.

\subsubsection{Extra features}

We trained another version of our model using additional per-segment features for a total of 49 input features: segment chord (start-to-end vector), chord length, segment arc length, segment mid-point, incoming and outgoing angle with the previous/next segments (encoded with continuous angle representation \cite{zhou2019continuity} as $(sin(\theta),cos(\theta))$), a unit vector from the panel center to the segment center, the length of that vector, and 5 equidistant tangent and curvature samples along the segment. The results, reported in Table \ref{table:auto-sewing-results} show no performance improvement, suggesting that all these secondary features are either irrelevant to the task or easy for the model to derive from the raw Bézier features.

\subsubsection{Parity matrix loss}

The parity matrices are extremely unbalanced, as most segment pairs do not belong to the same seam. We tried applying hard negative mining \cite{liu2016ssd,shrivastava2016training} to focus training on the most relevant signals. We used a 20:1 negative-to-positive ratio, always retaining the highest-scoring negative logits corresponding to negative targets, and we used equal positive and negative group weighting to balance the loss.

Let $\mathcal{P}$ and $\mathcal{N}$ denote the
positive and selected negative entries pooled across the batch.
We weight their mean losses equally:
\begin{equation*}
\mathcal{L}^*_{\mathrm{parity}}
=
-\frac{1}{2|\mathcal{P}|}
\sum_{u\in\mathcal{P}}\log\sigma(s_u)
-\frac{1}{2|\mathcal{N}|}
\sum_{u\in\mathcal{N}}\log(1-\sigma(s_u)),
\end{equation*}
where $s_u$ is a parity logit and $\sigma$ is the sigmoid function.

The results of a training run of our model with this modified loss are reported in Table \ref{table:auto-sewing-results}, showing a very similar performance compared to the standard binary cross entropy (BCE) loss proposed in $\S$\ref{sec:auto-sewing-loss}. Interestingly, training actually takes longer as performance plateaus after $\sim$1450 epochs, so we keep the simpler standard BCE loss.

\subsection{Automatic arrangement evaluation}

\begin{figure*}[h!]
    \centering
    \includegraphics[width=\linewidth]{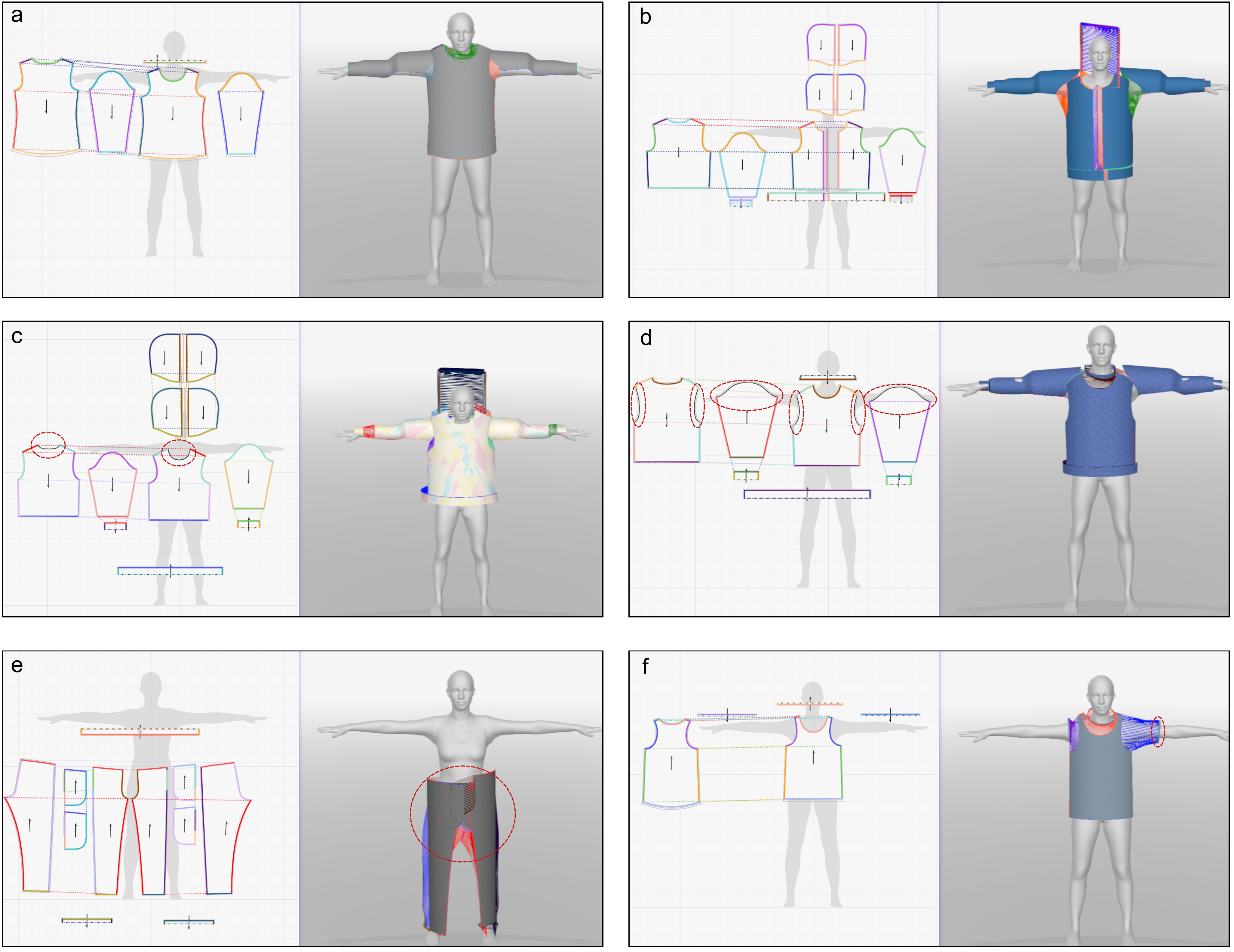}
    
    \caption{Examples of predicted sewing operations and panel arrangements in the CAD system. Each example shows the sewn 2D pattern (left) and the arrangement around the 3D body (right). Top row: correct sewing operations and arrangements. Middle row: missing seams. Bottom-row: incorrect arrangement assignments. Mistakes are circled in red.}

    \label{fig:result-samples}
\end{figure*}

\begin{table}
    \centering
    
    \caption{Results of the automatic arrangement experiments on the test set, listing percentage of correctly arranged panels and garments. Results for the ablation experiment are also reported.}
    
    \begin{tabular}{lrr}
        \hline
        Method & Garments (\%) & Panels (\%) \\
        \hline
        Ours            & \textbf{66.18} & \textbf{82.61} \\
        w/o Transformer     & 34.29 & 67.68 \\
        \hline
    \end{tabular}
     \label{table:auto-arrangement-results}
\end{table}

\begin{table}
    \centering
    
    \caption{Metrics corresponding to the evaluation outputs of the different automatic arrangement model heads. We report accuracy (acc.), precision (prec.), recall and F1 for body region and layer tasks, and Euclidean distance error for arrangement position.}
    
    \begin{tabular}{lrrrrr}
    \hline
    Task & Acc. & Prec. & Recall & F1 & E. distance (cm) \\
    \hline
    Body p. & 94.94 & 94.77 & 94.94 & 94.72 & - \\
    Layer   & 89.28 & 89.61 & 89.28 & 89.40 & - \\
    Pos.    & - & - & - & - & 3.76 \\
    \hline
\end{tabular}

    \label{table:arrangement-metrics}
\end{table}

For the quantitative evaluation of the automatic arrangement model performance on the test set, we consider a panel correctly arranged if it gets assigned to its ground truth body region and layer, and its arrangement position is within a 10 cm radius from the ground truth position (a reasonable tolerance given that the contribution of the arrangement position is marginal compared to that of the body region and layer assignments). We compute the percentage of panels correctly arranged over the total in the test set, and the percentage of garments with all panels correctly arranged. Table \ref{table:auto-arrangement-results} summarizes the results, with 66.18\% of perfectly arranged garments, and we report additional metrics in Table \ref{table:arrangement-metrics}. Looking at imperfectly arranged garments, we note that 73.4\% of them have fewer than 6 panels that would require manual adjustments. Some examples are shown in Figure \ref{fig:result-samples}.

\subsubsection{Impact of panel attention}

We evaluate the impact of inter-panel attention by training an alternative model without the Transformer encoder, directly routing $H_p$ to the model heads. The results, reported in Table \ref{table:auto-arrangement-results}, reveal a significant decrease in performance. A deeper look into the model errors reveals that most of the performance loss is due to a significant increase in mixing between first- and second-layer assignments, reasonable given the missing contextual information that helps decide whether a panel is part of an external or internal layer on a multilayer garment.
        
\subsection{Integration with the pipeline}

We have integrated our proposed models into existing apparel CAD software. When a new pattern is imported, our models can be run to sew and arrange the pattern. After that, a physics simulation runs to assemble and drape the garment on the target body. Finally, the True Seams technology \cite{rodriguez2022true} implemented in the CAD software, creates realistic seam and stitch geometry based on the topology and properties of the sewing operations. Figure \ref{fig:teaser} shows an instance of the full pipeline, going from imported pattern to draped and finished asset in a fully automatic manner for one of the test patterns for which all the arrangement and sewing operations are correctly inferred.

The percentage of garments in our test set that are both perfectly reconstructed and correctly arranged drops to 54.34\%, but since both methods integrate seamlessly with the existing seam creation and panel arranging tooling exposed by the CAD software, users can inspect and correct prediction errors using the existing editing tools. We discuss some examples, illustrated in Figure \ref{fig:result-samples}: top-row garments (a and b) have all sewing operations perfectly recovered and all panels correctly arranged, ready for simulation. The hoodie (c) is correctly arranged, but it is missing the seam that joins the hood to the body pieces, which must be created manually. In the case of the sweatshirt (d), all panels are correctly arranged, and most sewing operations are created, but the automatic sewing model predicted a spurious pair connecting the left and right armhole seams. This causes both seams to be bundled in the same connected component with disconnected order chains, getting discarded. Both seams must be created manually. The sewing operations of the sweatpants (e) are correctly recovered, but the front pieces of the pants are assigned to the front waist region instead of their respective front thigh regions, and one pocket panel was assigned to the incorrect layer. These three errors must be manually corrected before a successful simulation. The tank top (f) has all seams correctly created and all pieces are assigned to their correct body region and layer, but the arrangement position of one of them deviated $\sim$16 cm from its ground truth position. In this case, the simulation is robust to that mistake given the simplicity of the garment and no manual adjustment is needed.
\section{Conclusions}
In this work, we have introduced two novel architectures for automating the generation of sewing operations and the arrangement of an input raw vector 2D pattern. The proposed methods have been integrated into an apparel CAD system, enabling automatic generation of draped and finished 3D garments solely from an input 2D pattern.

Our proposed automatic sewing solution addresses multiple sewing operations at once, including many-to-many regular seams, facing seams, and hems, as well as seam configuration properties required to generate the finished virtual garment. Our proposed architecture operates solely on input vector segments, leveraging the Transformer architecture. Our experiments demonstrate that the model achieves its best performance without an explicit neighborhood attention mask or additional segment features, supporting our observation that implicit topology information is already present in the basic segment features and it is efficiently exploited by the attention mechanism.

The output of our automatic arrangement solution is adapted to the particularities of the tooling in the chosen CAD software, but the modular nature of the proposed architecture makes it easy to swap the output heads to adapt to different configurations.

\subsection{Further work}

Our experiments demonstrate progress towards automatic garment sewing and arrangement, but they also show how remaining prediction errors limit the reliability of fully automatic assembly. The proposed sewing model also excludes more complex, although arguably rare, overlapping constructions with three or more layers, and operations involving only part of a boundary segment.

Garments assembled by professional pattern makers provide valuable training data, but producing these annotations is labor-intensive and difficult to scale. Synthetic datasets generated using parametric models \cite{korosteleva2021generating,GarmentCodeData:2024} have greatly contributed to advances in this field. Extending these parametric models to support more complex overlapping constructions, partial-segment operations, and their construction properties could provide broader and more scalable supervision.



\bibliographystyle{IEEEtran}
\bibliography{references}

\end{document}